\def\year{2021}\relax
\documentclass[letterpaper]{article} 
\usepackage{aaai21}  
\usepackage{times}  
\usepackage{helvet} 
\usepackage{courier}  
\usepackage[hyphens]{url}  
\usepackage{graphicx} 
\usepackage{natbib}  
\usepackage{caption} 
\usepackage[utf8]{inputenc} 
\usepackage[T1]{fontenc}    
\usepackage{hyperref}       
\usepackage{url}            
\usepackage{booktabs}       
\usepackage{amsfonts}       
\usepackage{nicefrac}       
\usepackage{microtype}      

\usepackage{graphicx}
\usepackage{bm}
\usepackage{amsmath}
\usepackage{caption}
\usepackage{subcaption}
\usepackage{multirow}
\usepackage{xcolor}
\usepackage{adjustbox}
\usepackage[symbol]{footmisc}
\title{Customized Graph Neural Networks}
\title{Customized Graph Neural Networks}
\author {
    Yiqi Wang\footnotemark[1],\textsuperscript{\rm 1}
    Yao Ma\footnotemark[1],\textsuperscript{\rm 1}
    Wei Jin,\textsuperscript{\rm 1} \\
    Chaozhuo Li,\textsuperscript{\rm 2}
    Charu Aggarwal,\textsuperscript{\rm 3}
    Jiliang Tang\textsuperscript{\rm 1}\\
}
\affiliations {
    \textsuperscript{\rm 1} Michigan State University \\
    \textsuperscript{\rm 2} Microsoft Research Asia \\
    \textsuperscript{\rm 3} IBM T. J. Watson Research Center \\
    wangy206@msu.edu, mayao4@msu.edu, jinwei2@msu.edu, cli@microsoft.com, charu@us.ibm.com, tangjili@msu.edu
}

\begin{document}

\maketitle
\renewcommand{\thefootnote}{\fnsymbol{footnote}}
\footnotetext[1]{equal contribution}

\begin{abstract}
Recently, Graph Neural Networks (GNNs) have greatly advanced the task of graph classification. Typically, we first build a unified GNN model with graphs in a given training set and then use this unified model to predict labels of all the unseen graphs in the test set. However, graphs in the same dataset often have dramatically distinct structures, which indicates that a unified model may be sub-optimal given an individual graph. 
Therefore, in this paper, we aim to develop customized graph neural networks for graph classification. 
Specifically, we propose a novel customized graph neural network framework, i.e., Customized-GNN. Given a graph sample, Customized-GNN can generate a sample-specific model for this graph based on its structure. Meanwhile, the proposed framework is very general that can be applied to numerous existing graph neural network models. Comprehensive experiments on various graph classification benchmarks demonstrate the effectiveness of the proposed framework. 
\end{abstract}

\section{Introduction}\label{sec:introduction}

Graphs are natural representations for many real-world data such as social networks~\cite{yanardag2015deep,hamilton2017inductive,kipf2016semi,velivckovic2017graph}, biological networks~\cite{borgwardt2005protein,shervashidze2011weisfeiler} and chemical molecules~\cite{gilmer2017neural,rong2020self}. A crucial step to perform downstream tasks on graph data is to learn better representations. Deep neural networks have demonstrated great capabilities in representation learning for Euclidean data and thus have advanced numerous fields including speech recognition~\cite{nassif2019speech}, computer vision~\cite{he2016deep} and natural language processing~\cite{devlin2018bert}. However, they cannot be directly applied to graph-structured data since graphs have complex topological structures. Recently, graph neural networks (GNNs) have generalized deep neural networks to graph data. GNNs typically update node representations by transforming, propagating and aggregating node features across the graph. They have boosted the performance of many graph related tasks such as node classification~\cite{kipf2016semi,hamilton2017inductive}, link prediction~\cite{zhang2018anrl,gao2019graph,vashishth2020compositionbased}, and graph classification~\cite{ying2018hierarchical,ma2019graph,li2020graph,gao2021topology}. 

\begin{figure*}[ht]%

    \centering
    \subfloat[Node size distribution\label{fig:DD-distribution}]{{\includegraphics[width=0.24\linewidth]{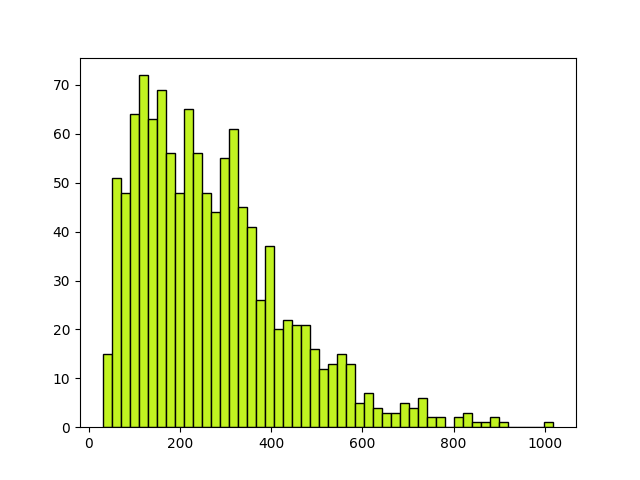}}}%
    \subfloat[A graph with 31 nodes\label{fig:DD-31-nodes}]{{\includegraphics[width=0.24\linewidth]{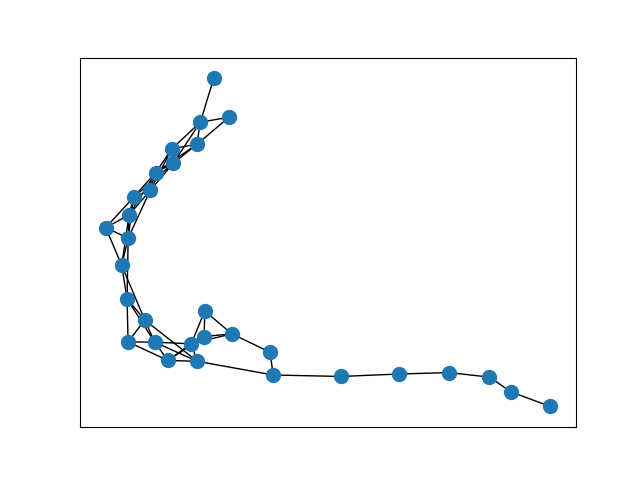} }}%
    \subfloat[A graph with 302 nodes \label{fig:DD-302-nodes}]{{\includegraphics[width=0.24\linewidth]{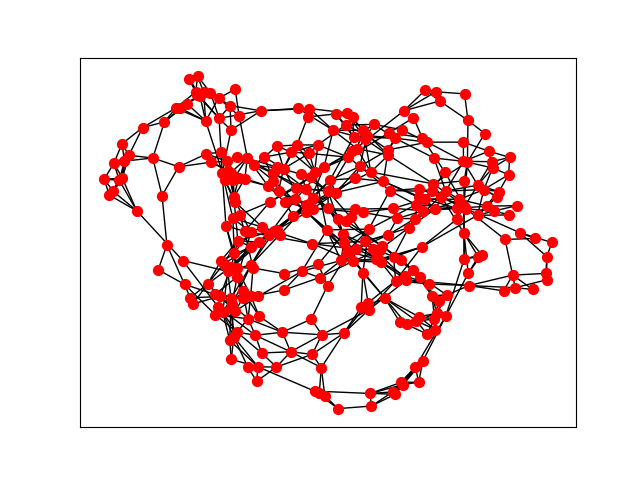}}}%
    \subfloat[Classification accuracy \label{fig:classification-acc}]{{\includegraphics[width=0.24\linewidth]{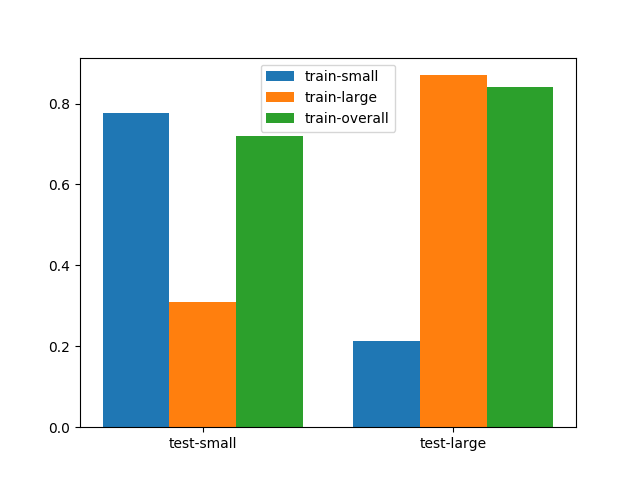} }}%
    \qquad

    \caption{An illustrative example of varied structural information and its impact on the performance of graph neural network based graph classification.}%

    \label{fig:DD-intro}
    \vspace{-0.1in}
\end{figure*}

Graph classification is one of the most important and prevalent graph related tasks~\cite{errica2019fair}, and in this work, we aim to advance graph neural networks for the graph classification task. 
There are numerous real-world applications for graph classification. 
For example, proteins can be denoted as graphs~\cite{dobson2003distinguishing} and the task to infer whether a protein functions as an enzyme or not can be regarded as a graph classification task; 
and it can also be applied to forecast Alzheimer's disease progression in which individual brains are represented as graphs~\cite{song2019graph}. 
Unlike data samples in classification tasks in other domains such as computer vision~\cite{rawat2017deep} and natural language processing~\cite{kowsari2019text}, graph samples in the graph classification task are described not only  by the input (node) features but their graph structures. Both the input node features and the graph structures play crucial roles in the graph classification tasks~\cite{ying2018hierarchical,ma2019graph,gao2021topology}.

In reality, graphs in the same data set can present significantly different structural properties.
Figure~\ref{fig:DD-distribution} demonstrates the distribution of graph size (i.e., the number of nodes) for protein graphs in the D\&D dataset~\cite{dobson2003distinguishing}, where the graph size varies dramatically from $30$ to $5,748$.
We further illustrate two graphs sampled from the D\&D dataset in Figures~\ref{fig:DD-31-nodes} and~\ref{fig:DD-302-nodes}, respectively. 
These two graphs present very different structural properties such as the number of nodes, graph shapes, and diameters.

The above investigations indicate that graphs in the same data set could have dramatically distinct structural properties. 
It naturally raises the question -- whether we should treat these graphs differently? 
To investigate this question, we take the graph size as the representative structure-property and demonstrate how it affects the graph classification performance. Specifically, we divide graphs from D\&D into two groups based on their graph size -- one for graphs with a small number of nodes and the other for graphs with a large number of nodes. Then, we split each group into a training set and a test set. Next, we train three GNN models \footnote[2] {The GNN model for graph classification uses GCN~\cite{kipf2016semi} as the filtering operation and maxpooling as the pooling operation.} based on three training sets -- the small one, the large one and the overall (a combination of the small and large one), separately. Then, we test their performance on the two test sets. The results are shown in the Figure~\ref{fig:classification-acc}. 
In the test set with small/large graph sizes, the model trained on the training set with the same graph size can significantly outperform the other two models. 
Investigations and consistent observations on more settings can be found in the Section ``Preliminary Data Analysis".

These investigations suggest that a unified model is not optimal for graphs with diverse structure properties and efforts are desired to consider the structure-property difference among graphs. Hence, in this paper, we aim to learn ``customized models'' for graphs with different structural properties. 
A natural way is to divide the dataset into different splits according to structure properties and train a model for each split. However, we face enormous challenges to achieve this goal in practice. First, there are potentially several structure properties (graph size, density, and etc.) affecting the performance, and we have no explicit knowledge about how the graphs should be split according to these properties. Second, dividing the dataset leads to small training sets for the splits, which may not be sufficient to train satisfactory models.
To address these challenges, we propose a novel graph neural network framework, Customized-GNN, for graph classification. The Customized-GNN framework is trained on \emph{all graphs in the given training set} (without splitting) and able to produce customized GNN models for each individual graph. Specifically, we design an adaptor, which is able to smoothly adjust a general GNN model to a specific one according to the structural properties of a given graph. The general GNN model and the adaptor are learned during the training stage simultaneously utilizing all graphs. 

Our major contributions are listed as follows: 1) We empirically observed that graphs in a given dataset could have dramatically distinct structural properties. Furthermore, it is not optimal to train a unified model for graphs with various structure properties for a graph classification task;
2) We propose a framework, Customized-GNN, which is able to generate a customized GNN model for each graph sample based on its structural properties. The proposed framework is general and can be directly applied to many existing graph neural network models;
3) We designed and conducted comprehensive experiments on numerous graph datasets from various domains to verify the effectiveness of the proposed framework.

\section{Preliminary Data Analysis}\label{sec:data_analysis}

In Figure~\ref{fig:DD-intro}, we have demonstrated that graphs in D\&D have varied properties, which affected the performance of GNNs for graph classification. In this section, we aim to further investigate this phenomenon by answering the following two questions -- (1) can the observations on D\&D be extended to other datasets? and (2) whether incorporating these properties into the models can facilitate the performance?

We choose four representative graph datasets from different domains for this study including \textbf{D\&D}~\cite{dobson2003distinguishing}, \textbf{ENZ}~\cite{shervashidze2011weisfeiler}, \textbf{PROT} ~\cite{borgwardt2005protein} and \textbf{RE-BI}~\cite{yanardag2015deep}.
We checked the properties such as node size and edge size. Similar to D\&D, graphs in all datasets present very diverse properties. More details about these datasets can be found in the {\bf Appendix A}. Following the same setting as D\&D, we divide each data into two groups according to the node size, i.e., large training and test (denoted as ``L-training" and ``L-test") and small training and test (indicated as ``S-training" and ``S-test"). 
The results are demonstrated in Table~\ref{table:nodesize}. 
From the table, we make consistent observations with these in D\&D -- models trained on one property group (e.g., L-training) cannot perform well on the other property group (e.g., S-test).

\begin{table}[!h]
\footnotesize
\centering
\caption{Graph classification accuracy on different node-size sets.}
\vspace{-0.1in}
\setlength{\tabcolsep}{0.6mm}{
\begin{tabular}{|c|ll|ll|ll|ll|ll|ll|}
\hline
               Accuracy ($\%$) & \multicolumn{2}{c|}{\textbf{D\&D}} & \multicolumn{2}{c|}{\textbf{ENZ}} & \multicolumn{2}{c|}{\textbf{PROT}} & \multicolumn{2}{c|}{\textbf{RE-BI}}\\ \hline
                         & S-test &L-test  & S-test &L-test  & S-test    &L-test & S-test   &L-test     \\\cline{1-9} 
S-training    & 66.2 & 55.7   & 45.0  & 20.0   & 76.5  &51.4  &88.6 &46.6         \\ \cline{1-9}
L-training & 47.2 & 77.8 & 27.0  & 39.0  & 45.6  & 78.6      &29.3 &83.1         \\\hline
\end{tabular}
\label{table:nodesize}
}

\end{table}

\begin{table}[]
\footnotesize
\setlength{\tabcolsep}{0.9mm}{
\centering
\caption{The classification accuracy of the models trained from four training sets in D\&D dataset and  Statistics for four training sets.}
\vspace{-0.1in}
\begin{tabular}{|c|c|c|c|c|c|c|}

\hline
\multirow{2}{*}{Training set} & \multirow{2}{*}{Node size range} & \multirow{2}{*}{\#Graphs}  &
\multicolumn{4}{c|}{Accuracy} \\ \cline{4-7} 
   &             &        & test 1 & test 2  & test 3 & test   \\ \hline

training 1  &[0,200]  & 369        & 76.1            & 41.2      & 26.7      & 50.9     \\ \hline
training 2  &[200,400]  & 392          & 43.5            &75.3      & 82.2      & 62.4        \\ \hline
training 3   &[400,2000]  & 180       &23.9           & 75.3       & 88.9      & 56.8          \\ \hline
training     &[0,2000]  & 941        & 64.1            & 61.9      & 75.6      & 66.2          \\ \hline
\end{tabular}
\label{tab:ac_on_diff_test}}
\vspace{-0.05in}

\end{table}

\begin{figure}[!t]
\begin{center}
{\includegraphics[width=1.05\linewidth]{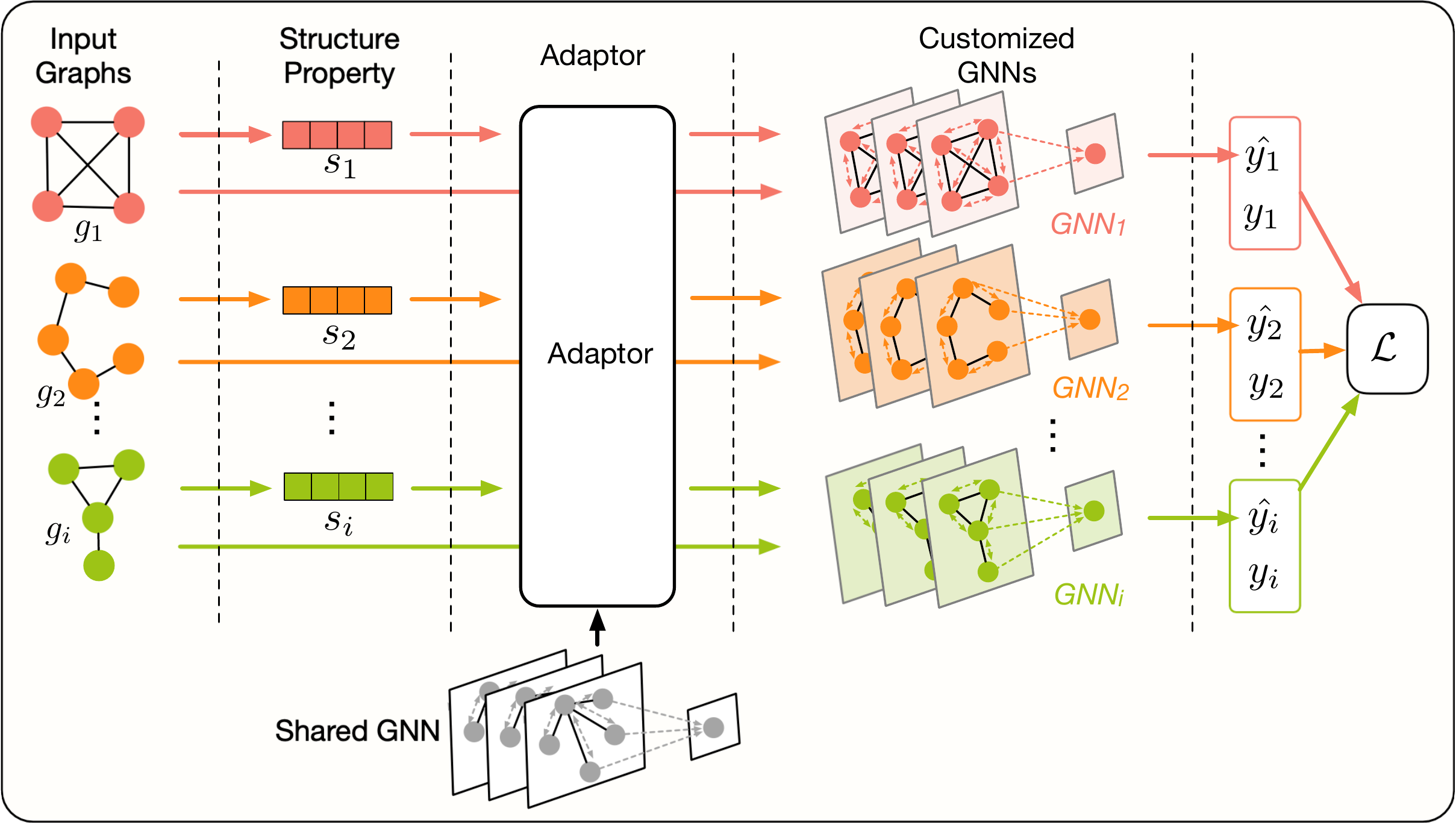}}
\vskip -0.1in
\end{center}

\caption{An overview of the proposed customized graph neural networks.}

\label{fig:model-frame}
\vspace{-0.15in}
\end{figure}
To answer the second question, we divide D\&D into several subsets based on the node size, and then divide each subset into a sub-training set ($80\%$) and a sub-test set ($20\%$). We train models on different sub-training sets separately, and then test their performance on all the sub-test sets. 
Specifically, we have trained four models on four different training sets from D\&D, which are {\it training 1}, {\it training 2}, {\it training 3} containing graph samples with node sizes from different ranges, and {\it training} which is the combination of {\it training 1}, {\it 2} and {\it 3}. 
Then, we test four models on four test sets, i.e., {\it test 1}, {\it test 2}, {\it test 3} and {\it test} which is the combination of {\it test 1}, {\it 2} and {\it 3}. Statistics about these training sets are summarized in Table~\ref{tab:ac_on_diff_test}.

The performance of four models on the test sets are illustrated in Table~\ref{tab:ac_on_diff_test}. 
We note that the model trained on a specific training set performs much better on the corresponding test set that shares the same node size range than the other test sets. This suggests the potential to incorporate the structure properties into the model training. 
In addition, though {\it training 1}, {\it training 2} and {\it training 3}  have much fewer training samples, the models trained on specific training sets can achieve better performance on the corresponding test sets compared to these trained from the entire training set (or {\it training}). 
This indicates that a unified graph neural network that is trained from the entire training set is not optimal for graphs with various structure properties in the test set.

{\bf Discussion.} Via the preliminary data analysis, we have established: (1) graphs in real-world data present distinct structure properties that tend to impact the graph classification performance of GNNs; and (2) incorporating the difference has the potential to boost the graph classification performance.  These observations lay the foundations of the model design in the next section.

\section{The Proposed Framework}\label{sec:model}

In this section, we introduce the proposed framework Customized-GNN that has been designed for graphs with inherently distinct structure properties. 
\begin{figure}[!t]
\begin{center}
{\includegraphics[width=1.1\linewidth]{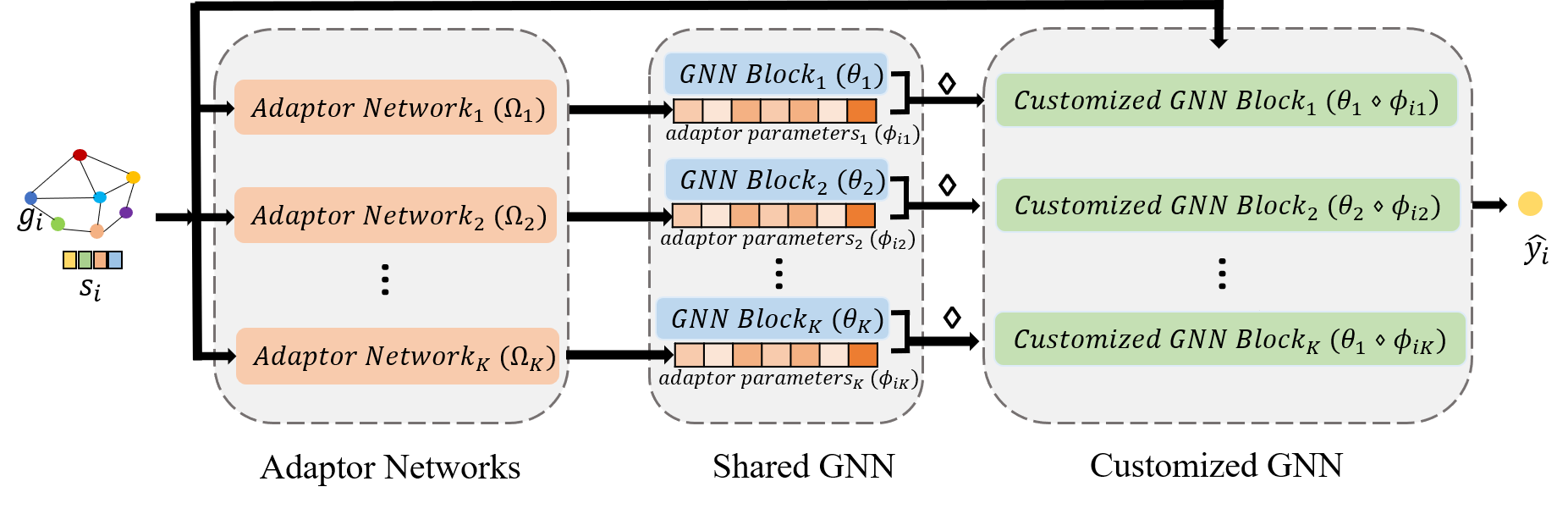}}
\vskip -0.1in
\end{center}

\caption{An overview of the GNN adaptation process.}

\label{fig:adaptor-frame}
\vspace{-0.1in}
\end{figure}
\subsection{The Overall Design}

As mentioned in earlier sections, graphs in real-world data inherently present distinct structural properties. Thus, we are desired to build distinct GNN models for them. To achieve this goal, we face tremendous challenges. First, we have no explicit knowledge about how the graph structure properties will influence graph neural network models. Second, if we separately train different models for graphs with different structure properties, we have to split the training sets for each model; as a consequence, the training data for each model could be very limited. For example, in the extreme case where each graph has unique graph structural properties, we only have one training sample for the corresponding model. Third, even if we can well train distinct GNN models for different graphs, during the test stage, for an unlabelled graph with unseen structural property, it is hard to decide which trained model we should adopt to make the prediction. In this work, we propose a customized graph neural network framework, i.e., Customized-GNN, which can tackle the aforementioned challenges simultaneously. 

An overview of the architecture of Customized-GNN is demonstrated in Figure~\ref{fig:model-frame}. The basic idea of Customized-GNN is -- it generates customized adaptor parameters for each graph sample $g_i$ via an adaptor network with the graph structure properties as input. These generated adaptor parameters are used to adapt a shared GNN model denoted as $GNN$ (this could be any GNN model that works for the graph classification task) to a model specific for the graph sample $g_i$. The adapted model $GNN_i$ incorporates the structure information of graph $g_i$, and thus, is customized for the graph sample $g_i$.

With the proposed Customized-GNN framework, the first challenge is handled, since the influence is implicitly modeled by the adaptor networks, which can customize the shared GNN model to a graph sample specific one. Furthermore, Customized-GNN can be trained on the entire training set without splitting it according to graphs' structure properties. This not only solves the second challenge but also ensures that the trained model can preserve common knowledge from the entire training set. The third challenge is also automatically addressed by the Customized-GNN framework. Given an unseen graph $g_j$, the Customized-GNN framework first takes its graph structure information as input and generates adaptor parameters. Then, these generated adaptor parameters can be used to customize the general GNN model to a customized one $GNN_j$ to predict the label of $g_j$.

Next, we introduce details about the adaptor network, the process of adapting a shared model to a specific one for a given graph, and the time complexity analysis of the proposed framework. Note that the training and test processes are similar to these of the traditional GNNs that can be found in the {\bf Appendix B}.

\subsection{The Adaptor Network}
The goal of the adaptor network is to generate the adaptor parameters for a given graph. From the preliminary data analysis, we have the intuition that the customized GNN model for a specific graph sample should be correlated to its structural properties. However, there is no explicit knowledge about how these structural properties influence graph neural network models. To model this implicit mapping function, we propose to utilize a powerful neural network to generate the model adaptor parameters from the observed structure information of a given sample.

In addition, graph neural networks often consist of several subsequent filtering and pooling layers, which can be viewed as different GNN blocks. For example, $K$ GNN blocks are shown in Figure~\ref{fig:adaptor-frame}. The graph structure properties of a given sample may have different influences on different GNN blocks. Hence, for each GNN block, we introduce one adaptor network to generate adaptor parameters for each block.

Specifically, we first extract a vector ${\bf s}_i$ to denote the structure information of a given graph ${g}_i$. We will discuss more details about ${\bf s}_i$ in the experiment section. As shown in the Figure~\ref{fig:adaptor-frame}, the adaptor networks take the structure information ${\bf s}_i$ as input and generate the adaptation parameters for each block. In the case where there are $K$ blocks in the graph neural network, we have $K$ independent adaptor networks corresponding to the $K$ blocks. Note that these adaptor networks share the same input ${\bf s}_i$ while their outputs are different. Specifically, the adaptor network for the $j$-th block can be expressed as follows:
\begin{equation}
    \bm{\phi}_{ij}= {h_j}(\bm{s}_i; \Omega_{j}), j = 1,\dots, K,\label{eq:adap_para_gen}
\end{equation}
where $\Omega_j$ denotes the parameters of the $j$-th adaptor network and $\bm{\phi}_{ij}$ denotes its output, which will be used to adapt the $j$-th learning block. The adaptor network ${h}_j$ can be modeled using any functions. In this work, we utilize feed-forward neural networks due to their strong capability.
According to the universal approximation theorem ~\cite{hornik1991approximation}, a feed-forward neural
network can approximate any nonlinear functions. 
For convenience, we summarize the process of the $K$ adaptor networks with $s_i$ as input below:
\begin{align}
    \Phi_i = H({\bf s}_i; {\bf \Omega}_H ),
    \label{eq:adapator_para}
\end{align}
where $\Phi_i$ contains the generated adaptation parameters of all the GNN blocks for graph $g_i$ and ${\bf \Omega}_H$ denotes the parameters of the $K$ adaptor networks.

\subsection{ The Adapted Graph Neural Network}\label{the_adapted_model}

Any existing graph neural network model can be adapted by the Customized-GNN framework to generate sample-specific models based on the structure information. Therefore, we first generally introduce the GNN model for graph classification and describe how to adapt it given a specific sample. Then, we illustrate how to adapt specific GNN models. 

\subsubsection{A General Adapted Framework}

A typical GNN framework for graph classification contains two types of layers, i.e., the filtering layer and the pooling layer. The filtering layer takes the graph structure and node representations as input and generates refined node representations as output. The pooling layer takes graph structure and node representations as input to produce a coarsened graph with a new graph and new node representations. A general GNN framework for graph classification contains $K_p$ pooling layers, each of which follows $K_f$ stacking filtering layers. Hence, there are $K=K_p*K_f$ learning blocks in the GNN framework. A graph-level representation can be obtained from these layers that can be further utilized to perform the prediction. Given a graph sample $g_j$, we need to adapt each of the $K$ layers according to its adaptor parameters generated from the adaptor network. Via this process, we can generate a GNN model $GNN_j$ specific to $g_j$.

Without loss of generality, when introducing a filtering layer or a pooling layer, we use an adjacency matrix ${\bf A} \in \mathbb{R}^{n\times n}$ and node representations ${\bf X}\in \mathbb{R}^{n\times d}$ to denote the input of these layers where $n$ is the number of nodes and $d$ is the dimension of node features. Then, the operation of a filtering layer can be described as follows:
\begin{align}
    {\bf X}_{new} = f({\bf A}, {\bf X} ; {\bf \theta}_f)  
\end{align}
where $\theta_f$ denotes the parameters in the filtering layer and ${\bf X}_{new} \in \mathbb{R}^{n\times d_{new}}$ denotes the refined node representations with dimension $d_{new}$ generated by the filtering layer. Assuming $\phi_f$ is the corresponding adaptor parameters for this filtering layer, we adapt the model parameter $\theta_f$ of this filtering layer as follows:
\begin{align}
\theta_f^m =  \theta_f \diamond \phi_f,    
\end{align}
where $\theta_f^m$ is the adapted model parameter that has the same dimension as the original model parameter $\theta_f$; and $\diamond$ is the adaptation operator. The adaption operator can have various designs, which can be determined according to the specific GNN model. We will provide the details of the adaptation operator when we introduce concrete examples in the following subsections. Then, with the adapted model parameters, we can define the adapted filtering layer as follows:
\begin{align}
    {\bf X}_{new} = f({\bf A}, {\bf X} ; {\bf \theta}_f\diamond \phi_f ). 
\end{align}

On the other hand, the process of a pooling layer can be described as follows:
\begin{align}
    {\bf A}_{new}, {\bf X}_{new} = p({\bf A}, {\bf X}; {\bf \theta}_p),
\end{align}
where $\theta_p$ denotes the parameters of the pooling layer, ${\bf A}_{new} \in \mathbb{R}^{n_{new} \times n_{new}}$ with $n_{new} <n$ is the adjacency matrix for the newly generated coarsened graph and ${\bf X}_{new}\in \mathbb{R}^{n_{new} \times d_{new}}$ is the learned node representations for the coarsened graph. Similarly, we adapt the model parameters of the pooling layer as follows:
\begin{align}
\theta_p^m =  \theta_p \diamond \phi_p,    
\end{align}
which leads to the following adapted pooling layer:
\begin{align}
    {\bf A}_{new}, {\bf X}_{new} = p({\bf A}, {\bf X}; {\bf \theta}_p\diamond \phi_p),
\end{align}
where $\phi_p$ is the adaptation parameters generated by the adaptor network for this pooling layer.

For convenience, we summarize a general GNN model as \newline     $GNN(\cdot |\Theta_{GNN} )$, where $\Theta_{GNN}$ is the parameters in all GNN blocks(i.e., $\theta_f, \theta_p$ in all filtering and pooling layers). Then, for a graph sample $g_i$, we can adapt the GNN model  $GNN(\cdot |\Theta_{GNN} )$ to a customized model for $g_i$ denoted as $GNN(\cdot |\Theta_{GNN} \Diamond \Phi_i)$. Note that, as shown in Eq.~\eqref{eq:adapator_para}, $\Phi_i$ contains adaptation parameters of all GNN blocks for a graph sample $g_i$. The adaptation operations in all GNN blocks (including filtering and pooling layers) are summarized in $\Theta_{GNN} \Diamond \Phi_i$. There are numerous GNN models designed for graph classification~\cite{gao2019graph,ranjan2019asap,ma2019graph,Yuan2020StructPool:}. The proposed framework can be applied to the majority of these models, i.e., these models all can serve as the $GNN(\cdot |\Theta_{GNN} )$ model mentioned above. In this work, we focus on three representative GNN models including GCN~\cite{kipf2016semi}, DiffPool~\cite{ying2018hierarchical} and gPool~\cite{gao2019graph}. We would like to leave the investigations of other GNN models as one future work. Next, we will give details on how to adapt GCN and  DiffPool since gPool follows a similar adaptation process.

\subsubsection{Adapted GCN: Customized-GCN}\label{sec:adapted_gcn}
Graph Convolutional Network (GCN)~\cite{kipf2016semi} is originally proposed for semi-supervised node classification task. The filtering layer in GCN is defined as follows:
\begin{equation}
{\bf X}_{new} = f({\bf A}, {\bf X} ; {\bf \theta}_f )=  \sigma(\tilde{\bf D}^{-\frac{1}{2}}\tilde{\bf A}\tilde{\bf D}^{-\frac{1}{2}}{\bf X}{\bf W}), 
\label{eq: filter_gcn}
\end{equation}
where $\tilde{\bf A} = {\bf A} + {\bf I}$ represents the adjacency matrix with self-loops, $\tilde{\bf D} = \sum_{j}\tilde{\bf A}_{ij}$ is the diagonal degree matrix of $\tilde{\bf A}$ and ${\bf W}\in \mathbb{R}^{d\times d_{new}}$ denotes the trainable weight matrix in filtering layer and $\sigma(\cdot)$ is a nonlinear activation function. With the adaptation parameter $\phi_f$ for this corresponding filtering layer, the adapted filtering layer can be represented as follows:
\begin{equation}
{\bf X}_{new} = f({\bf A}, {\bf X} ; {\bf \theta}_f\diamond \phi_f )=  \sigma(\tilde{\bf D}^{-\frac{1}{2}}\tilde{\bf A}\tilde{\bf D}^{-\frac{1}{2}}{\bf X}({\bf W}\diamond \phi_f)). 
\label{eq:filter_gcn_mo}
\end{equation}
Specifically, we adopt FiLM~\cite{perez2018film} as the adaption operator. In this case, the dimension of the adaptor parameter is $2d$, i.e., $\phi_f \in \mathbb{R}^{2d}$. We split $\phi_f$ into two parts $\gamma_f\in \mathbb{R}^d$ and $\beta_f\in \mathbb{R}^d$ and then the adaptation operation can be expressed as follows:
\begin{align}
    {\bf W} \diamond \phi_f = ({\bf W} \odot br(\gamma_f,d_{new}) ) + br(\beta_f, d_{new}),
    \label{eq:film}
\end{align}
where $br({\bf a}, k)$ is a broadcasting function that repeats $k$ times for the vector ${\bf a}$; hence, $br(\gamma_f,d_{new})\in \mathbb{R}^{d\times d_{new}}$ and $br(\beta_f,d_{new})\in \mathbb{R}^{d\times d_{new}}$ have the same shape as ${\bf W}$ and $\odot$ denotes the element-wise multiplication between two matrices.  

To utilize GCN for graph classification, we introduce a node-wise max pooling layer to generate graph representation from the node representations as follows:
\begin{align}
    {\bf x}_G = p({\bf A}, {\bf X} ; \theta_p) = max({\bf X}),
\end{align}
where ${\bf x}_G \in \mathbb{R}^{1 \times d_{new}}$ denotes the graph-level representation and $max()$ takes the maximum over all the nodes. Note that the max-pooling operation does not involve learnable parameters and thus no adaptation is needed for it. We refer to an adapted GCN framework as Customized-GCN.

\subsubsection{Adapted diffpool: Customized-DiffPool}\label{sec:adapted_diff}
DiffPool is a hierarchical graph level representation learning method for graph classification~\cite{ying2018hierarchical}. The filtering layer in DiffPool is the same as Eq.~\eqref{eq: filter_gcn} and its corresponding adapted version is shown in Eq.~\eqref{eq:filter_gcn_mo}. Its pooling layer is defined as follows:
\begin{align}
    &{\bf S} = softmax( f_a({\bf A}, {\bf X}; \theta_{f_a})), \label{eq: pool_filter}\\
    &{\bf X}_{new}  = {\bf S}^T {\bf Z}, \label{eq:feature}\\
    &{\bf A}_{new} =  {\bf S}^T{\bf A} {\bf S}, \label{eq:structure} 
\end{align}
where $f_a$ is a filtering layer embedded in the pooling layer, ${\bf S}\in \mathbb{R}^{n\times n_{new}}$ is a soft-assignment matrix, which softly assigns each node into a supernode to generate a coarsened graph. Specifically, the structure and the node representations for the coarsened graph are generated by Eq.~\eqref{eq:structure} and Eq.~\eqref{eq:feature} respectively, where ${\bf Z} \in \mathbb{R}^{n \times d_{new}}$ is the output of the filtering layers. To adapt the pooling layer, we only need to adapt Eq.~\eqref{eq: pool_filter}, which follows the same way as introduced in Eq.~\eqref{eq:filter_gcn_mo} as it is also a filtering layer. We refer to the adapted diffpool model as Customized-DiffPool.

\subsection{Time Complexity Analysis}
In this subsection, we analyze the additional time required to calculate the adaptation parameters and perform the adaptation. Specifically, we use the FiLM adaptation operator,
as an example for the adaptor network. For convenience, the dimension of the output node features in all layers is assumed to be the same $d$. The dimension of the output of the adaptation network $\phi_f$ is $2d$. Furthermore, we assume that the input of the adaptation network, i.e., the graph property information ${\bf s}_i$ is with dimension $s$. Then, the time complexity to generate the adaptation parameters for a single block using Eq.~\eqref{eq:adap_para_gen} is $O(2d\cdot s)=O(d\cdot s)$. Furthermore, the time required to adapt the parameters for a single block with Eq.~\eqref{eq:film} is $O(d^2)$. Hence, for graph neural networks with $K$ learning blocks, the time complexity to calculate the adaptation parameters and perform the adaptation for all learning blocks is $O(K\cdot d\cdot s + K\cdot d^2)$. Note that, the time complexity of a single filtering operation in Eq.~\eqref{eq: filter_gcn} is $O(m\cdot d + n \cdot d^2)$ where $m$ denotes the number of edges while $n$ is the number of nodes.
Therefore, the total time complexity for $K$ learning blocks without adaptation is $O(K\cdot m\cdot d +K \cdot n \cdot d^2)$. Furthermore, $s$ is typically small (much smaller than $m$); hence, the additional time complexity introduced by the adaptation operation is rather small.
\section{Experiment}\label{sec:experiments}

In this section, we conducted comprehensive experiments to verify the effectiveness of the proposed Customized-GNN framework. We first describe the experimental settings. Then, we evaluate the performance of the framework by comparing original GCN, DiffPool and gPool with the adapted GCN, DiffPool, gPool models by the Customized-GNN framework. Next, we analyze the importance of different components in the adaptor operator. Finally, we conduct case studies to further facilitate our understanding of the proposed method.

\begin{table*}
\footnotesize
 \caption{Comparisons of graph classification performance in terms of accuracy.} 
  \centering

  \begin{tabular}{c|ccccccc}
  \bottomrule

  \multirow{2}{*}{Accuracy ($\%$)}&\multicolumn{7}{c}{Datasets} \\ \cline{2-8}
  
                  & \textbf{COLLAB}   & \textbf{ENZ}  & \textbf{PROT}     & \textbf{DD}       & \textbf{RE-BI}    & \textbf{RE-5K}       & \textbf{NCI109} \\ \bottomrule
GCN              & 67.9$\pm$1.4 & 50.4$\pm$3.0 & 77.0$\pm$2.3 & 79.3$\pm$5.3 & 82.6$\pm$4.9 & 50.7$\pm$1.3 & 74.9$\pm$2.7    \\
Concat-GCN       & 68.4$\pm$1.4 & 52.5$\pm$5.1 & 78.4$\pm$1.9 & 77.6$\pm$3.2 & 80.7$\pm$3.5 & 50.7$\pm$1.0 & 75.6$\pm$1.2 \\
Multi-GCN-2    &68.3$\pm$1.4 & 47.0$\pm$1.8 &  79.5$\pm$1.3 & 77.1$\pm$2.4 & 80.6$\pm$3.5 & 50.3$\pm$1.9 & 74.2$\pm$1.9 \\
Multi-GCN-3  &67.0$\pm$1.4 &44.6$\pm$5.4 &79.9$\pm$2.2  & 76.7$\pm$3.5  & 77.5$\pm$7.3 &48.5$\pm$2.1 &75.8 $\pm$1.5\\
Customized-GCN  & 71.3$\pm$1.0 & 55.4$\pm$4.6 & 78.8$\pm$3.2 & 79.6$\pm$3.9 & 91.5$\pm$1.6 & 53.3$\pm$1.3 & 76.7$\pm$1.1 \\ \bottomrule
DiffPool     & 70.6$\pm$1.2 & 57.9$\pm$2.5 & 78.6$\pm$2.5 & 81.5$\pm$4.1 & 89.6$\pm$1.1 & 56.2$\pm$1.1& 77.5$\pm$0.7 \\
Concat-Diff & 70.7$\pm$0.7 & 60.0$\pm$1.7 & 77.7$\pm$2.5 & 81.3$\pm$2.9 & 91.1$\pm$1.7 & 54.9$\pm$1.4 & 78.0$\pm$0.5 \\
Multi-Diff-2   &70.7$\pm$0.6 &56.3$\pm$1.3  &80.0$\pm$1.2 &79.3$\pm$2.9 &89.9$\pm$2.5 &53.7$\pm$0.6 &76.8$\pm$0.7 \\
Multi-Diff-3  &70.8$\pm$1.1 &52.5$\pm$0.8 &80.9$\pm$1.7 &80.6$\pm$2.3 &88.8$\pm$0.7 &53.4$\pm$2.4 &78.5$\pm$1.2 \\
Customized-DiffPool  &73.6$\pm$0.5 &57.9$\pm$7.2 &78.6$\pm$2.9 &80.6$\pm$2.6 &95.1$\pm$1.6 &55.8$\pm$1.1 & 78.2$\pm$0.9 \\ \bottomrule
gPool   &69.4$\pm$2.2 &53.8$\pm$3.2 &77.3$\pm$3.0 &78.9$\pm$5.5 &88.9$\pm$1.6 &51.3$\pm$0.6 &77.1$\pm$1.2 \\
Concat-gPool &69.7$\pm$0.5 &57.1$\pm$1.8 &79.1$\pm$2.2 &78.0$\pm$2.6 &88.5$\pm$1.3 &50.9$\pm$2.2 &76.3$\pm$0.7 \\ 
Multi-gPool-2 &69.0$\pm$1.9 &50.8$\pm$5.1  &79.7$\pm$1.0 &79.5$\pm$2.7 &84.0$\pm$3.2 &49.3$\pm$2.3 &73.5$\pm$2.1 \\
Multi-gPool-3 &68.9$\pm$1.6 &46.2$\pm$3.2  &80.6$\pm$0.8  &80.0$\pm$3.6  &83.1$\pm$4.5 &48.9$\pm$1.8 &75.2$\pm$1.9 \\
Customized-gPool  & 72.3$\pm$1.0 & 62.9$\pm$3.6 & 80.6$\pm$1.6 & 80.0$\pm$3.1 &91.1$\pm$0.7 &53.3$\pm$1.4 &76.5$\pm$1.9 \\ \bottomrule

  \end{tabular}

 \label{table:graph_classfication}
\end{table*}

\subsection{Experimental Settings}
We carried out graph classification tasks on seven datasets. More details about these datasets can be found in the {\bf Appendix A}. Next, we describe the baselines. In the Section ``The Proposed Framework" , we apply the proposed framework to adapt three graph neural networks models: a basic graph convolutional network (GCN)~\cite{kipf2016semi}, and two SOTA graph classification models  DiffPool~\cite{ying2018hierarchical} and gPool~\cite{gao2019graph}, respectively. The corresponding adapted versions are Customized-GCN, Customized-DiffPool and Customized-gPool, respectively. {\it Our evaluation purpose is if the proposed framework can boost the performance of existing models by adapting them to their corresponding customized versions.} Thus, (1) to validate the effectiveness of the proposed model, we compare Customized-GCN, Customized-DiffPool, Customized-gPool with GCN, DiffPool and gPool; and (2) we have not chosen models in~\cite{ranjan2019asap,ma2019graph,Yuan2020StructPool:} as baselines here but the proposed framework can be directly applied to adapt them as well. Note that in this work, we construct a set of simple structural features ${\bf s}_i$ of $g_i$ such as the number of nodes, the number of edges and the graph density; however, it is flexible to include other complex features by the proposed framework. Furthermore, we create baselines to directly concatenate the graph structure properties ${\bf s}_i$ to the output graph embedding of the GCN, DiffPool and gPool model. Correspondingly, we call these three methods as Concat-GCN, Concat-Diff and Concat-gPool. In addition, we develop baseline methods, Multi-GCN, Multi-Diff and Multi-gPool. They learn multiple graph convolutional networks for graph samples with different structural information. \textbf{Multi-GCN (or Multi-Diff, Multi-gPool)} consists of several GCN (or DiffPool, gPool) models trained from different subsets of the training dataset. Specifically, we first cluster data samples from the training set into different training subsets based on the graph structural information. Then we train different models from different training subsets. During the test phase, given a test graph sample, we first assign it to one cluster with the smallest Euclidean distance between its graph structural information and the centroid of the training cluster. Then, we choose the model trained on the cluster for prediction. In this experiment, we set the number of clusters to $2$ and $3$, and denote the corresponding frameworks as Multi-GCN-2 (or Multi-Diff-2, Multi-gPool-2) and Multi-GCN-3 (or Multi-Diff-3, Multi-gPool-3). More details of these baselines can be found in the {\bf Appendix C}.

\begin{table}

 \caption{Adaptability study. (Note here Cust-X denotes Customized-X)}
 \vspace{-0.1in}
 \begin{adjustbox}{width=0.5\textwidth}

  \centering
  \begin{tabular}{c|cc|cc|cc}
  \hline
  \multirow{2}{*}{Accuracy(\%)}&\multicolumn{6}{c}{Methods} \\ \cline{2-7}
  
                 &Cust-GCN& GCN   &Cust-DiffPool    & DiffPool &Cust-gPool    & gPool \\ \hline
\textbf{ENZ}     &  22.2  &  20.5 &25.6  &  22.2 & 35.0&  24.8            \\
\textbf{RE-BI}   &  70.2  &   50.4  & 78.6 &  52.7 & 80.0 & 59.9                \\
\hline
  \end{tabular}
  
 \end{adjustbox}
 \label{table:adaptability}
  \vspace{-0.2in}
\end{table}

\begin{figure}[ht]%
\vspace{-0.05in}
    \centering
    \subfloat[Model Embeddings \label{fig:case-study-model}]{{\includegraphics[width=0.5\linewidth]{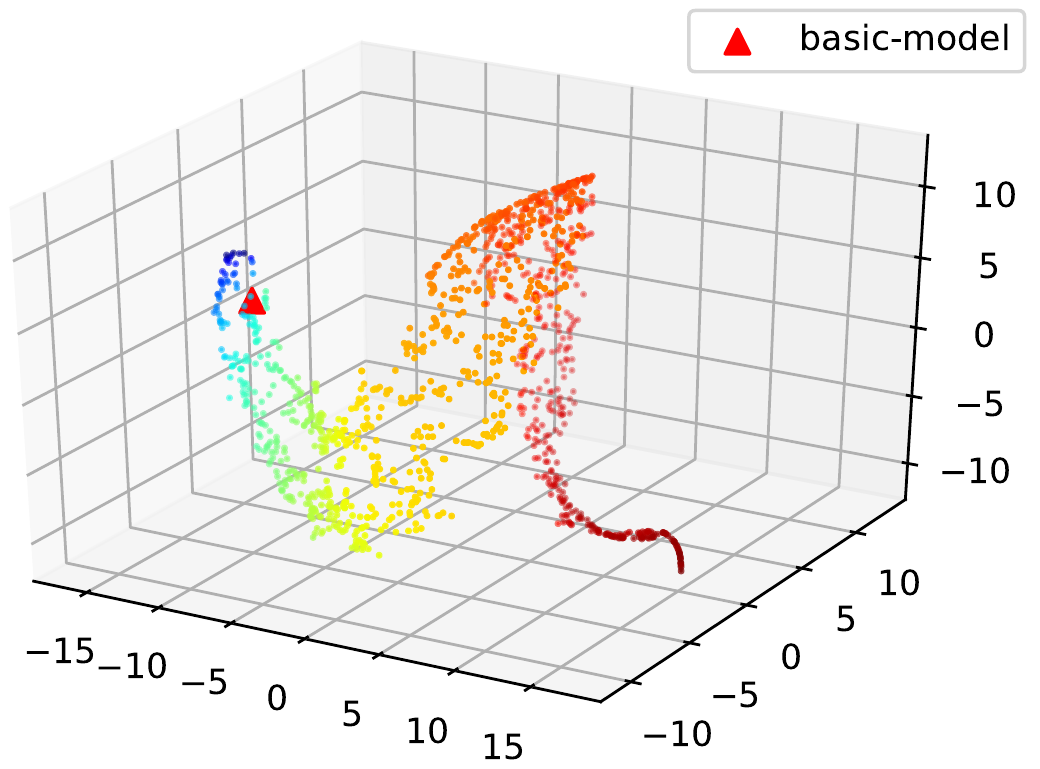}}}
    \subfloat[Sample demonstration \label{fig:case-study-example}]{{\includegraphics[width=0.5\linewidth]{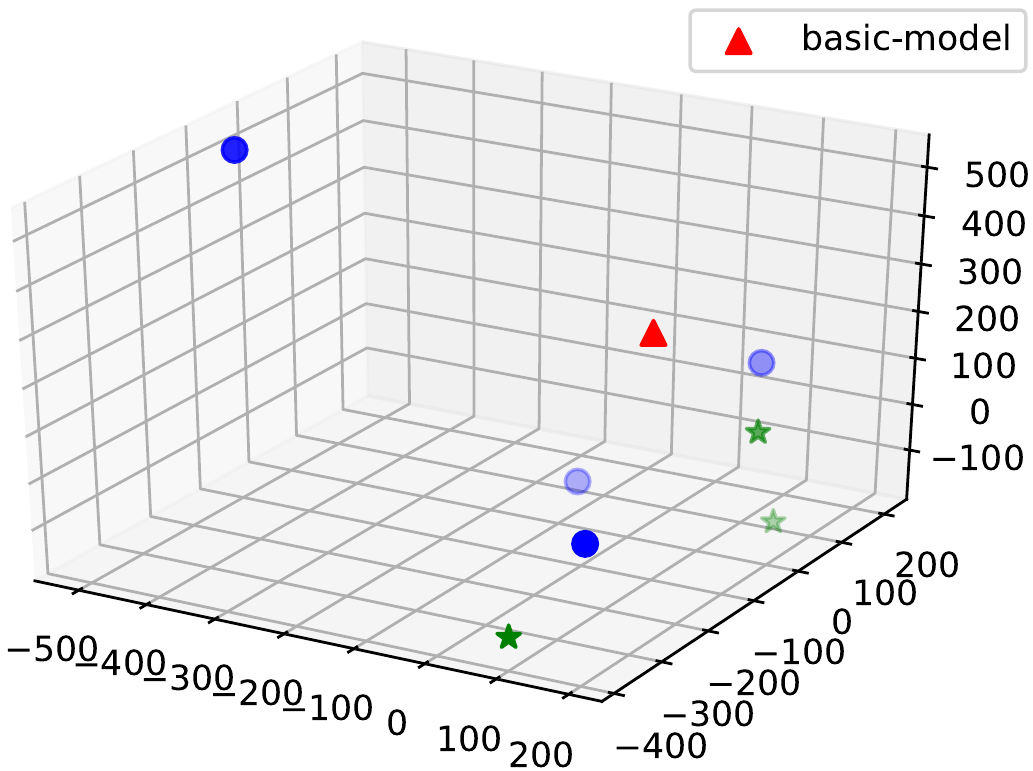} }}
    \caption{Case Study. (a) depicts the model embeddings and (b) demonstrates model embeddings for graphs that are mistakenly classified by GCN, but correctly classified by Customized GCN.}%
    \label{fig:case-study}
    \vspace{-0.2in}
\end{figure}
\vspace{-0.1in}

\subsection{Graph Classification Performance Comparison}

In this subsection, we first perform the comparison following the traditional setting. To further demonstrate the advantage of the proposed frameworks, we show their adaptability when the properties of test graphs are different from these of training graphs. Following the setting in~\cite{ying2018hierarchical}, for each graph dataset, we randomly shuffle the dataset and then split $90\%$ of the data as the training set and the remaining $10\%$ as test set.  We train all the models on the training set and evaluate their performance on the test set with accuracy as the measure. We repeat this process with different data shuffling and initialization seeds for $4$ times and report the average performance and standard variance. More implementation details can be found in the {\bf Appendix D}. 

The results are shown in Table~\ref{table:graph_classfication}. We notice that the Concat- and the Multi- version of the GNN models can, in some cases, achieve comparable or even better performance than their corresponding original versions. This indicates that utilizing the graph structure properties has the potential to help improve the model performance. However, the performance of these variants is not so stable across different datasets, which means that these simple methods are not suitable for all datasets. For example, the Concat- versions may work well on datasets where the label is directly related with the graph structure properties but fail on those datasets where graph structure properties have more implicit impact on the labels. On the other hand, the performance of the Multi-version of the GNN models is heavily dependent on how the data is split into different groups. It is not practical to find good splits manually. Furthermore, simply training different models for different graphs can lead to unsatisfactory performance because less training data is available for each model. In contrast, our proposed Customized models learn sample-wise adaptation, which automatically finds suitable models for different data samples according to their graph structure properties. Compared with the original GCN, DiffPool and gPool, the corresponding Customized models achieve better performance in most of the datasets. This demonstrates that the sample-wise adaptation performed by the Customized-GNN framework can boost the performance of GNN frameworks.

\begin{table}
\caption{Ablation study.}
 \vspace{-0.1in}
\begin{adjustbox}{width = 0.5\textwidth}

  \centering
  \begin{tabular}{c|ccccccc}
  \hline
  \multirow{2}{*}{Accuracy (\%)}&\multicolumn{7}{c}{Datasets} \\
                    & \textbf{COLLAB}   & \textbf{ENZ}  & \textbf{PROT}     & \textbf{DD}       & \textbf{RE-BI}    & \textbf{RE-5K}       & \textbf{NCI109} \\ \bottomrule
GCN             & 69.9   &51.8  & 76.6 & 77.2  &81.9   &50.4    & 75.7        \\
Customized-GCN$_{\gamma}$ & 70.8   & 52.3  & 77.6  & 78.1 &85.2  &51.7    & 76.0     \\
Customized-GCN$_{\beta}$  &71.2   &54.0  &77.9   & 78.0 & 88.8 & 51.9   &77.1            \\\hline
Customized-GCN    &73.2   & 55.9  & 77.9 & 79.3  & 90.4 &52.9   & 77.1        \\\hline
  \end{tabular} 
  \end{adjustbox}
  \label{table:ablation}
   \vspace{-0.2in}
\end{table}

{\bf Adaptability Study.} To further show the adaptability of the proposed framework to new graphs with different structures, we order graphs according to their node sizes in non-decreasing order. Then, we use the first $80\%$ of the data as training set and the remaining $20\%$ as test set. The purpose of this setting is to simulate the case where structures of graphs in the test set are different from those in the training set. We only show the results on the \textbf{ENZ} and \textbf{RE-BI} datasets in Table~\ref{table:adaptability}, since observation from other datasets are consistent. We note that (1) GCN, DiffPool and gPool cannot work properly in this setting; and (2) the customized frameworks perform much better under this setting. These results demonstrate the ability of the learned Customized-GNNs to adapt GNNs to graphs with new properties.

\subsection{Ablation Study} 

In this subsection, we investigate the effectiveness of different components in the adaptor operator in Eq.~\eqref{eq:film} used in our model. Specifically, we want to investigate whether $\gamma_f$ and $\beta_f$ play important roles in the adaptor operator by defining the variants of Customized-GCN -- \textbf{Customized-GCN$_{\gamma}$}: It is a variant of the adaptor operator with only element-wise multiplication operation where instead of Eq.~\eqref{eq:film}, the adaptation process is now expressed as: $ {\bf W} \diamond \phi_f = ({\bf W} \odot br(\gamma_f,d_{new}) ) $; and \textbf{Customized-GCN$_{\beta}$}: It is a variant of the adaptor operator with only element-wise addition operation where instead of Eq.~\eqref{eq:film}, the adaptation process is now: $ {\bf W} \diamond \phi_f = {\bf W} +  br(\beta_f,d_{new})$. Following the previous experimental setting, we compared Customized-GCN with its variants.
The results are presented in Table~\ref{table:ablation}. We observe that both \textbf{Customized-GCN$_{\gamma}$} and \textbf{Customized-GCN$_{\beta}$} can outperform the original GCN model. It indicates that both terms with $\gamma$ and $\beta$ are effective for the adaptation and utilizing either one of them can already adapt the original model in a reasonable manner. We also note that the Customized-GCN model outperforms both \textbf{Customized-GCN$_{\gamma}$} and \textbf{Customized-GCN$_{\beta}$} on most of the datasets. It demonstrates that the adaption effects of the term with $\gamma$ and $\beta$ are complementary to each other and combining them together can further enhance the performance.

\vspace{-0.05in}
\subsection{Case Study}

To further illustrate the effectiveness of the proposed framework, we conducted case studies on D\&D. First, we visualize the distribution of embeddings of sample-specific model parameters for different graph samples. Specifically, we take the parameters of the first filtering layer of each sample-specific Customized-GCN framework and then utilize t-sne~\cite{maaten2008visualizing} to project these parameters to $3$-dimensional embeddings. We visualize these $3$-d embeddings in the form of a scatter plot as shown in Figure~\ref{fig:case-study-model}. Note that in this figure, the red triangle denotes the embedding of the parameters (i.e. ${\bf W}$) of the original GCN model (the one before adaptation). For each point in the figure, we use color to represent the scale of values in terms of node size. Specifically, a deeper red color indicates a larger value, while a deeper blue color indicates a smaller value. We make some observations from Figure~\ref{fig:case-study-model} . First, the proposed Customized-GCN framework indeed generates distinct models for different graph samples that are different from the original model. Second, the points with similar colors stay closely with each other, which means that graphs with similar structural information share similar models. In addition, in Figure~\ref{fig:case-study-example}, we illustrate the sample-specific model parameters for seven samples with different numbers of nodes. They are misclassified by the original GCN model but correctly classified by the proposed Customized-GCN framework. It is obvious that Customized-GCN has generated seven different GCN models for these graph samples, each of which can successfully predict the label for the corresponding sample. More case study results can be found in the {\bf Appendix E}.

\section{Related work}\label{sec:related work}

Graph neural networks have advanced a wide variety of tasks.
including node classification~\cite{kipf2016semi,hamilton2017inductive}, link prediction~\cite{gao2019graph,vashishth2020compositionbased} and graph classification~\cite{ying2018hierarchical,ma2019graph,li2020graph,gao2021topology}.
In the task of graph classification, one of the most important steps is to get a good graph-level representation. A straightforward way is to directly summarize the graph representation by globally combining the node representations~\cite{duvenaud2015convolutional}. Recently, there are some works investigating learning hierarchical graph representations by leveraging deterministic graph clustering algorithms~\cite{defferrard2016convolutional,fey2018splinecnn}. There also exist end-to-end models aiming at learning hierarchical graph representations, such as DiffPool~\cite{ying2018hierarchical}. MuchGNN~\cite{zhou2019multi} proposed to learn a set of graph channels at each layer to shrink the graph hierarchically. Furthermore, some methods~\cite{gao2019graph,gao2021topology} propose principles to select the most important $k$ nodes to form a coarsened graph in each layer. 
EigenPooling~\cite{ma2019graph} is based on graph Fourier transform and is able to capture the local structural information. In~\cite{Yuan2020StructPool:}, conditional random fields (CRF) are used to design the pooling operation.

\section{Conclusion}

In this paper, we propose a general graph neural network framework, Customized-GNN, to deal with graphs that have various graph structure properties. Comprehensive experiments demonstrated that the Customized-GNN framework can effectively adapt both flat  and hierarchical GNNs to enhance their performance. Future research directions include better modeling the adaptor networks, considering more complex properties, and adapting more existing graph neural networks models.  

\bibliography{sample-base}


\end{document}